# Linguistic Characteristics of AI-Generated Text: A Survey

**Luka Terčon**
Faculty of Arts, University of Ljubljana
Faculty of Computer and Information Science, University of Ljubljana

**Kaja Dobrovoljc**
Faculty of Arts, University of Ljubljana
Jožef Stefan Institute

Large language models (LLMs) are solidifying their position in the modern world as effective tools for the automatic generation of text. Their use is quickly becoming commonplace in fields such as education, healthcare, and scientific research. There is a growing need to study the linguistic features present in AI-generated text, as the increasing presence of such texts has profound implications in various disciplines such as corpus linguistics, computational linguistics, and natural language processing. Many observations have already been made, however a broader synthesis of the findings made so far is required to provide a better understanding of the topic. The present survey paper aims to provide such a synthesis of extant research. We categorize the existing works along several dimensions, including the levels of linguistic description, the models included, the genres analyzed, the languages analyzed, and the approach to prompting. Additionally, the same scheme is used to present the findings made so far and expose the current trends followed by researchers. Among the most-often reported findings is the observation that AI-generated text is more likely to contain a more formal and impersonal style, signaled by the increased presence of nouns, determiners, and adpositions and the lower reliance on adjectives and adverbs. AI-generated text is also more likely to feature a lower lexical diversity, a smaller vocabulary size, and repetitive text. Current research, however, remains heavily concentrated on English data and mostly on text generated by the GPT model family, highlighting the need for broader cross-linguistic and cross-model investigation. In most cases authors also fail to address the issue of prompt sensitivity, leaving much room for future studies that employ multiple prompt wordings in the text generation phase.

## 1 Introduction

In recent years, the use of Large Language Models (LLMs), such as ChatGPT (OpenAI 2025), has seen a fast increase particularly in fields such as education, healthcare, and scientific research (Gasparini et al. 2024; Kasneci et al. 2023). Texts generated by such artificial intelligence tools are becoming increasingly widespread, with their presence reported in various forms on different online platforms (Liang et al. 2024; Matsui 2024; Thompson et al. 2024). The language of such texts can appear very authentic, to the extent that humans often perform quite poorly when trying to determine if a text was written by a human or a large language model (Casal and Kessler 2023). However, various extant studies have found that texts generated by large language models differ in terms of certain linguistic characteristics from texts written by humans. For instance, AI-generated texts are less lexically diverse, syntactically more complex, and contain more nominalization than human-written texts (André et al. 2023; Herbold et al. 2023; Liu et al. 2023; Muñoz-Ortiz et al. 2024).

A new text type with its own specific set of linguistic characteristics is thus beginning to emerge, and with the fast growth of AI-generated text on the internet and in other media it is becoming crucial to systematically study these characteristics. Some authors have already brought into attention the issue of how this increase in AI-generated content will affect the language produced by human authors (Matsui 2024; Liang et al. 2024). To better understand this impact, further analysis of the characteristics of AI-generated text compared to human-written text is required. This is especially relevant in the domain of corpus linguistics, where the usual course of analysis is to inspect large quantities of authentic human texts. The rising use of automatic text generation tools increasingly blends together AI-generated and human-written content, raising the question of whether authentic human language use can still be commonly assumed. Additionally, the analysis of the linguistic features that characterize AI-generated texts is also beneficial for the disciplines of natural language processing and computational linguistics, where such studies can aid in the development of new methods of automatic AI-generated content detection and other ways of countering the malicious use of LLMs (Mindner et al. 2023; Petukhova et al. 2024; Shah et al. 2023; Zaitsu and Jin 2023).

Although many studies have examined the linguistic characteristics of AI-generated text, the findings remain scattered and unsynthesized. While existing reviews mainly target other questions, such as detecting generated text (Dhaini et al. 2023; Wu et al. 2025) or investigating the level of awareness of underlying linguistic constraints exhibited by LMMs (Chang and Bergen 2024), they do not provide a systematic account of broader linguistic patterns observed when comparing AI-generated and human-generated text.

To address this gap, this work aims to provide a focused overview of the findings that have been made in relation to the linguistic characteristics of texts generated by LLMs in comparison to human-written texts. We put special emphasis on what can be found in the end result – the generated text – rather than the internal functioning of the models. By providing a synthesis of the findings presented in such linguistic studies, a clearer picture can be constructed of the inherent linguistic style present in the language generated by modern LLMs. However, despite the fact that several studies have presented compelling findings about the linguistic features of generated text, it may not be the case that all text produced by LLMs tends to consistently exhibit the same linguistic characteristics in all situations. Instead, LLM output may vary depending on the language of the text, its genre, the level of linguistic description in question, etc. To address this, in addition to a qualitative examination of the extant studies, the present survey also includes a quantitative overview, in which the works are classified according to several criteria, thus helping to illustrate which viewpoints and methodological issues have so far been considered.

In line with the above, we formulate two main research questions, which will guide our survey:

1. What are the main methodological tendencies of current research on the linguistic characteristics of AI-generated-text in comparison to human-written texts?
2. What are the common linguistic characteristics of AI-generated text reported in several scientific studies compared to human-written texts?

Henceforth, we refer to text that is produced by large language models as AI-Generated Text (AIGT), while text produced by human authors is referred to as Human-Written Text (HWT).

In Section 2 the method of obtaining relevant scientific articles is described, along with the method of classifying the obtained articles along several criteria. Section 3 continues with a general quantitative overview of the categorization of each article, followed by an introduction to the main linguistic findings in Section 4 and an analysis of the main findings in relation to several unique methodological considerations in Section 5. The findings are then analyzed and assessed for common points in a general discussion in Section 6. Section 7 concludes the survey paper.

## 2 Method

In this section, we explain the details of the research methods we employed. Our study consisted of three separate phases: in the first phase we collected all relevant literature. In the second phase we conducted a quantitative analysis of the obtained work through which we addressed our first research question. In the last phase, we organized the results of the analyzed studies into a connected whole, thus providing a summary of the most important findings that concern the topic at hand and addressing our second research question. In the following we explain the details of the literature collection phase (Section 2.1) and the quantitative analysis phase (Section 2.2).

### 2.1 Literature Collection

The websites ResearchGate[1] and Google Scholar[2] were employed to search for scientific works related to the topic. Keywords such as "AI generated text", "automatically generated text", "linguistic characteristics", and "language use" were used to obtain scientific works, which were subsequently inspected by one of the authors and included in the survey if they were deemed relevant. An article was considered relevant if it focused on an LLM that is based on the transformer architecture (Vaswani et al. 2017) and contained concrete findings regarding the unique linguistic characteristics of AIGT in comparison to HWT. Although the articles obtained are of diverse types, they can all be classified into two general categories with regard to their study objective: the first category is composed of studies that explicitly set out to uncover the linguistic characteristics of AIGT by analyzing the differences between AIGT and HWT. The second category, on the other hand, consists of studies that describe the development and evaluation of an AIGT detection system that uses linguistic features to classify text. These detection studies also include an analysis of which linguistic features prove most useful for their detection systems, thus providing valuable insight into the typical linguistic characteristics that set apart AIGT from HWT. Using this method a total of 44 articles were obtained.

### 2.2 Classification Categories

In the second phase of this survey, all the relevant obtained works were classified according to five different criteria: the levels of linguistic description studied, models used to generate the text, genres of text analyzed, languages analyzed, and approach to prompting. Categorization according to these criteria make it simpler to observe the main overall methodological trends that studies of this topic are currently following.

---

[1] https://www.researchgate.net/
[2] https://scholar.google.com/

For the levels of linguistic description criterion, we decided to categorize the analyzed studies in a way that is consistent with the traditional lexicon-grammar approach to language description. Namely, we focus on three different levels of linguistic analysis: lexicon, grammar, and other. The lexical level includes investigations of the typical vocabulary as well as the distribution of various groups of lexical items. Studies which looked at whether certain types of punctuation are characteristic of AIGT are also considered part of the lexical level of description, as well as studies that examine repetition patterns. The grammatical level includes observations that take into account linguistic notions such as the morphological structure of words, the syntactic makeup of sentences, and part-of-speech patterns in AIGT. The category labeled Other is used for studies whose methodologies also include approaches that do not fit neatly into the other two levels. Examples of such methodologies include, for instance, sentiment analysis and references to named entities.

When reporting on the specific type of large language model(s) used for their study, authors differ considerably in the degree of detail provided. Some, for example, provide a very specific description of the model version used (e.g. "ggml-gpt4all-j-v1.3-groovy" (Shah et al. 2023)), while some are much more vague. For the sake of simplicity, we classify studies using broader model family categories. These cover models that are generally referred to by a common name and design architecture (such as Llama 3 and BLOOM), but may include various models fine-tuned on different training datasets. This approach avoids excessive granularity in the reporting of the results, while still highlighting the commonalities between models that belong to the same family.

For the types of genres analyzed in different studies, we focus on the most commonly generated text genres and again include a special Other category, as the range of different AIGT genres used is quite extensive.

We also categorize the studies according to what languages the authors generate their AIGT in as well as the approaches they take to prompting their LLMs. Prompt design is an important aspect which any study examining the output of generative AI models should take into account. The concept of prompt engineering has gained much attention lately and special guides have been formulated to help researchers and other users obtain a more desirable output through refined prompt design techniques (Giray 2023; Meskó 2023). The sensitivity of LLM output to the form of prompt provided has been shown to be so great that different prompt wordings might even significantly change model rankings for certain evaluation tasks (Mizrahi et al. 2024; Voronov et al. 2024). To assess the level of awareness regarding prompt sensitivity, we categorize the studies based on whether they obtain their AIGT data using a single prompt or more than one concrete prompt wording. We also report if the study does not specify the employed prompt at all.

A complete table with the categorization of each analyzed study is provided in Appendix A.

# 3 Quantitative Overview

Figures 1-5 present a numerical overview of the classification of articles obtained using our method.

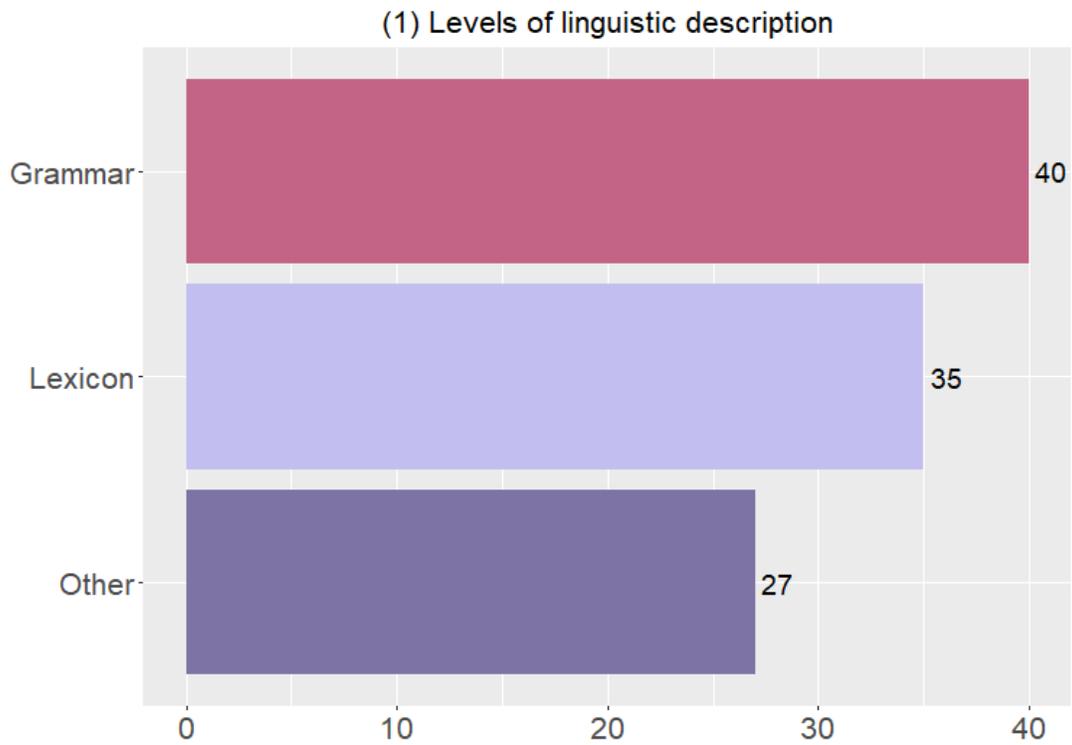

Figure 1: Number of obtained articles that use methods pertaining to each level of linguistic description

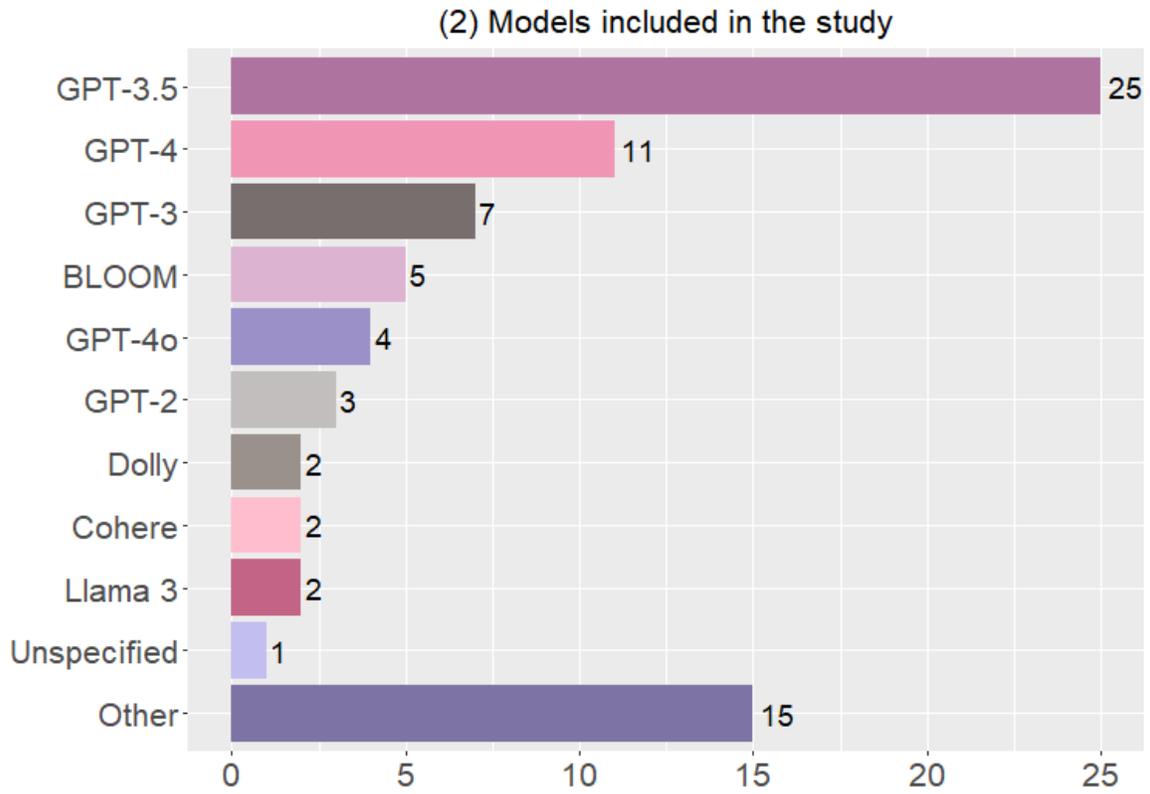
Figure 2: Number of obtained articles that use a certain large language model

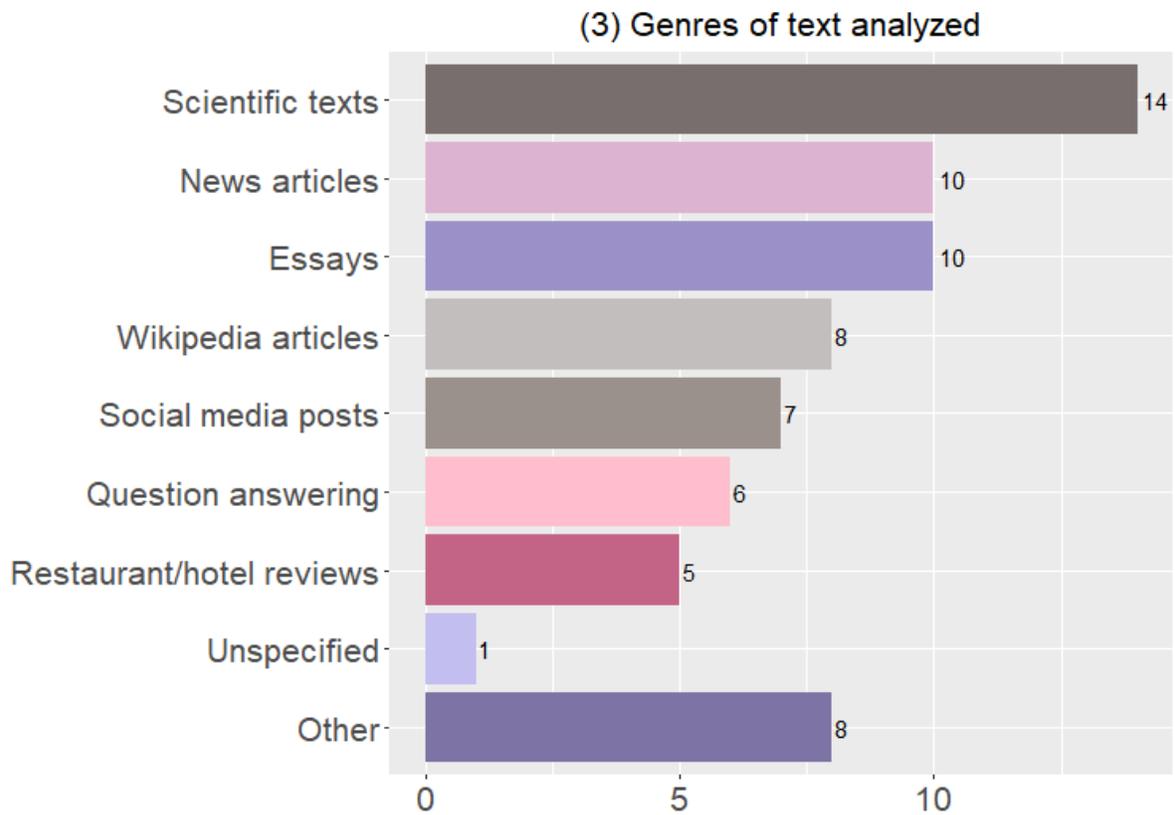

Figure 3: Number of obtained articles which analyze a specific genre of generated texts

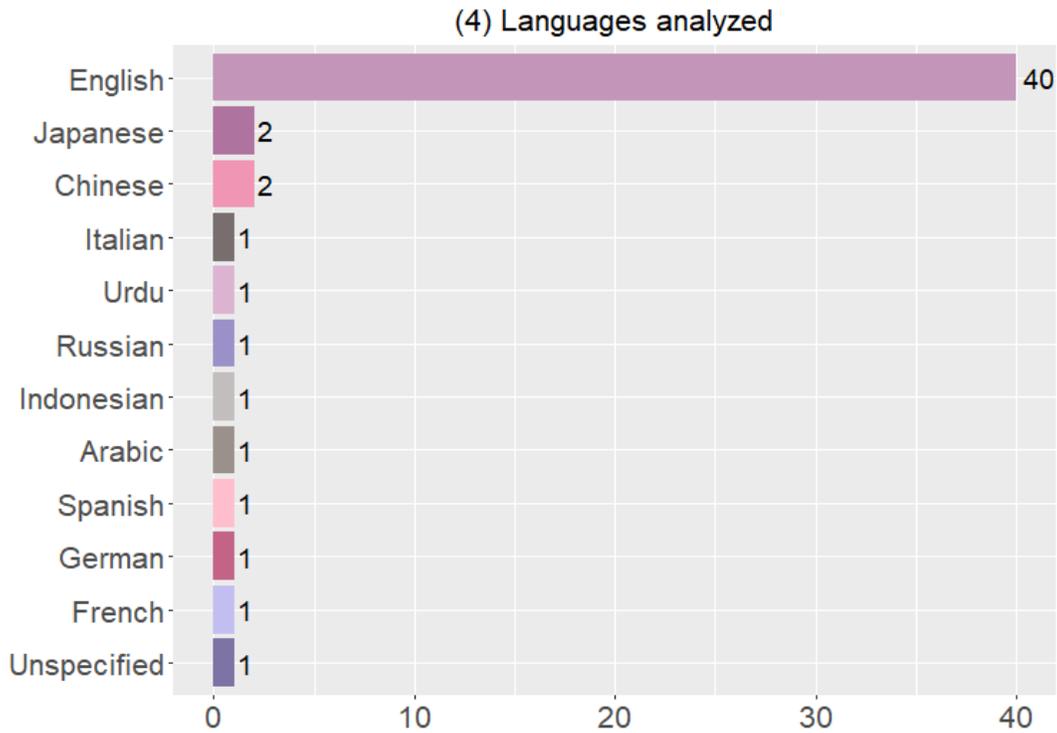

Figure 4: Number of obtained articles that analyze AIGT in a specific language

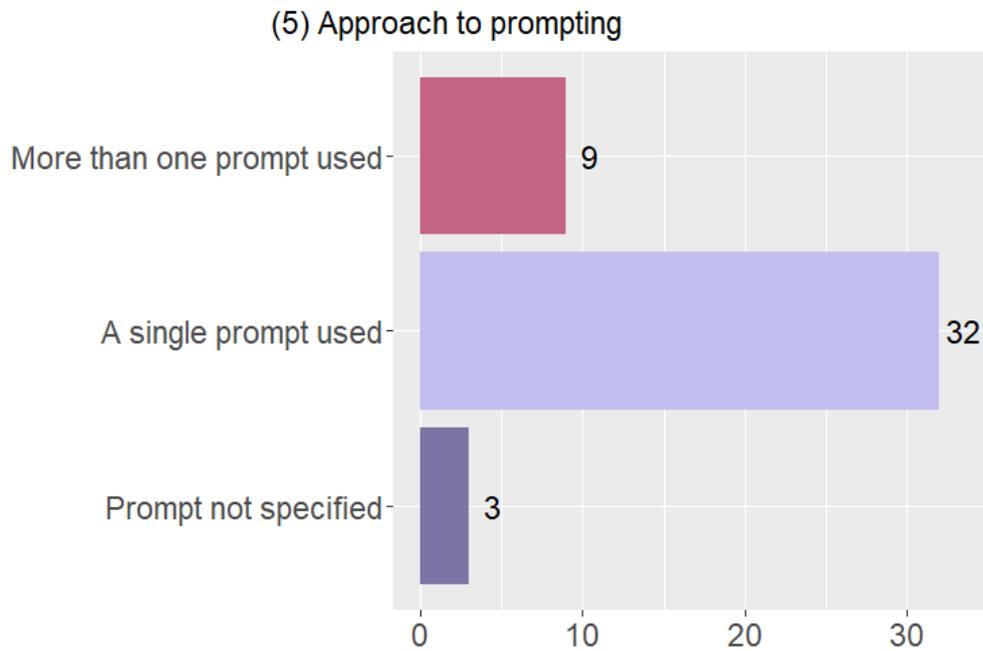

Figure 5: Number of obtained articles that use a specific approach to prompting

Figure 1 shows the number of obtained articles with methods pertaining to each level of linguistic description. The most often analyzed level of linguistic description appears to be grammar, which is addressed in 40 articles, closely followed by the lexical level with 35 articles, and the Other category with 27.

Figure 2 shows the categorization of each study based on the models employed. By far the most commonly used model appears to be GPT-3.5, with 25 works (approx. 57%) using it in their study. This most likely stems from the fact that the model was the most easily accessible and most widely known LLM at the time of writing for most studies. It is followed by GPT-4 and GPT-3, two other models developed by OpenAI. The most often studied open models (as opposed to the models maintained by OpenAI whose designs are not openly accessible to the public) are models from the multilingual BLOOM family. These are followed by other families such as GPT-4o, GPT-2, Dolly, and Cohere. The Other category is composed of a total of 15 models which are mentioned in only one study. In Figure 2 this group was combined into a single category to preserve space. These 15 models are OLMo, Mistral, Llama 2, OpenAI o1, Alpaca, Llama 1, Orca-Mini, GPT4All-J, Pythia, GPT-Neo, Qwen2.5, BingAI, Gemini, Bard, and one model based on the original Transformer. In one article, the authors specify only that they used ChatGPT for their analysis, without any further detail. We make the assumption that these authors ran their experiments on GPT-3.5, the most basic freely accessible ChatGPT model at time of publication of that article. One study did not specify the model type used in their experiments.

Figure 3 presents an overview of the most common genres of AIGT analyzed by the obtained articles. Scientific texts were the most common genre, investigated by 14 studies (approx. 32%). They are followed by news articles and essays, both with 10 studies. Other prominent genres of text include Wikipedia articles, social media posts, answers to questions provided in the prompt, and reviews of hotels and restaurants. One study left the genre of AIGT unspecified. The Other category included genres which were not mentioned in analyses very often and spanned a wide assortment of text types including, to name a few, simulation of spontaneous human conversations, email communication, explanations of technical terms, and web text.[3]

Figure 4 presents the most common languages of AIGT. We identified eleven different languages in total that the included articles cover as the language of the AIGT. The articles disproportionately often looked at linguistic characteristics of AIGT generated in English, as this language is studied in 40 articles (which corresponds to 91%). The difference between English and other languages is quite striking, especially when considering that Japanese and Chinese – the next two most commonly used languages – each appear in only 2 articles. The authors of one article do not explicitly state which language the texts were generated in. Since we were not able to discern this information from the article, we labeled the language as Unspecified.

The last figure, Figure 5, shows a categorization of the studies based on their approach to prompting. Regarding the prompting strategy employed, 9 (approx. 20%) out of the 44 papers report using more than one type of prompt to obtain AI-generated data for their experiments. In addition, 3 papers do not specify the prompt used to obtain AIGT.

---

[3] With web text we refer to a number of studies which used texts scraped from different websites as reference and prompted LLMs to produce comparable texts.

# 4 Linguistic Features

In this section we present the main findings of our obtained articles, organized according to the three levels of linguistic description used in the quantitative analysis, namely lexicon (Section 4.1), grammar (Section 4.2), and other (Section 4.3). This grouping enables us to present the findings in a concise and organized manner that is familiar to any linguist.

## 4.1 Lexicon

On the lexical level, linguistic studies of AIGT focus on several different notions, such as the vocabulary used, lexical diversity, repetition, and word length.

### 4.1.1 Vocabulary

A lot of attention has been devoted to specific English vocabulary items which are more commonly used by LLMs than human authors. Researchers have identified both content and function words that are specific to either type of text. Content words more likely to be found in AIGT than HWT include those that refer to ambiguous groups of people, such as *others* and *researchers,* and some other rare expressions like *stand out feature, incredibly polite, waitstaff,* and *knowledgeable,* while the verb *say* and words that are connected to specific feelings, such as *feels like, felt sick,* and the verb *hate,* are used less commonly in AIGT (Desaire et al. 2023; Kim and Desaire 2024; Mitrović et al. 2023). Sensing verbs, like *read*, *look*, and *hear* are also less common in AIGT (Wan 2024). As for function words, AIGT is less likely to contain the conjunctions *however, but, although, because, if, that, or, when, as,* the pronouns *this, I, they, my, it, us,* the prepositions *to, as, before, about, since,* the modal verbs *will, would, might, could,* and the adverbs *up, out, off*. On the other hand AIGT is more likely to contain the modal verb *can*, the conjunction *and*, the prepositions *by, with, for,* and the personal pronoun *their* (Desaire et al. 2023; Fujiwara 2024; Kim and Desaire 2024). In addition, aggressive and rude expressions are usually not found in AIGT (Mitrović et al., 2023).

### 4.1.2 Lexical Diversity

Lexical Diversity can be measured using a number of methods which usually provide information about how varied a vocabulary is present in some text. We find that, among the articles we analyzed, the authors almost universally report that AIGT is much less lexically diverse than HWT (André et al. 2023; Culda et al. 2025; Guo et al. 2023; Liao et al., 2023; Liu et al. 2023; Muñoz-Ortiz et al. 2024; Seals and Shalin 2023). Relatedly, Yildiz Durak et al. (2025) find that HWT features a considerably higher number of unique words. Only two studies deviate from this. Both claim that texts generated by LLMs score higher on measures of lexical diversity (Zhang and Crostwaithe 2025; Zindela 2023), though Zindela (2023) adds that the diversity of function words is lower in AIGT – a trend which is also described by André et al. (2023). When describing the development of their AIGT detection system, Shah et al. (2023) also report that their best performing systems were highly dependent on lexical diversity scores.

### 4.1.3 Repetition

Repetition is generally more commonly found in AIGT as human authors tend to avoid repeating the same words and expressions (Simon et al. 2023; Yanagita et al. 2024). AIGT also often exhibits repetitive emoticon use compared to HWT (Arcenal et al. 2024).

### 4.1.4 Word Length

André et al. (2023) discover that AIGT often contains longer tokens compared to HWT. However, Yildiz Durak et al. (2025) find no notable difference in the length of words between AIGT and HWT. Cai et al. (2023) find that LLMs do not choose a shorter word over a longer one when the context is predictable – a tendency that human speakers have been shown to prefer.

### 4.1.5 Miscellaneous

Idiomatic expressions are reportedly less commonly found in texts produced by LLMs than those written by human authors (Simon et al. 2023). AIGT also often contains non-existent words. When analyzing AI-generated clinical vignette texts in Japanese, Yanagita et al. (2024) detect names of diseases and drugs that are nonsensical or non-existent. With regard to the types of punctuation marks used, AIGT tends to contain a less varied set of punctuation marks, with commas and periods accounting for most cases (Simon et al. 2023). AIGT contains less commas, question marks, dashes, parentheses, semicolons, and colons compared to HWT, while having more single quotes and periods (Desaire et al. 2023; Simon et al. 2023). Several authors who describe the development of an AIGT detection system also report that the performance of their systems greatly improved when the input included counts of words belonging to a specific lexical category (Mikros et al. 2023; Mindner et al. 2023; Nguyen et al. 2023).

## 4.2 Grammar

On the grammatical level, studies analyze diverse aspects of AIGT, including notions such as syntactic complexity, sentence length, the distribution of parts-of-speech and syntactic structures, the ordering of sentence constituents, and n-gram patterns.

### 4.2.1 Syntactic Complexity

The authors of the analyzed studies most often measure syntactic complexity using metrics based on dependency distance,[4] though some authors also use other less common metrics such as dependency tree depth and mean length of T-unit (Herbold et al. 2023; Liu et al. 2023). A total of five studies find that there are differences in syntactic complexity between AIGT and HWT. Among these, four agree that AIGT is usually more syntactically complex (Culda et al. 2025; Herbold et al. 2023; Liu et al. 2023; Nkhobo and Chaka 2023), while Seals and Shalin (2023) only mention that there is a difference, without elaborating on the relationship any further.

Three studies focus on the dependency lengths of specific syntactic structures. Of these three, two reveal that AIGT contains shorter dependency lengths with conjunct relations (Guo et al. 2023; Liu et al. 2023), while Guo et al. (2023) also find that punctuation dependency lengths are much longer. Liao et al. (2023) observe that nominal subjects have shorter dependency lengths.

---
[4] Most commonly, a higher average dependency distance is connected to higher syntactic complexity.

### 4.2.2 Sentence Length

Zindela (2023) and Desaire et al. (2023) note that sentence lengths in AIGT tend to vary much less compared to HWT. With the average length of sentences, things are a little less clear, as some authors find notable differences between AIGT and HWT (Culda et al. 2025; Guo et al. 2023; Liu et al. 2023; Zindela 2023) and others claim that there is no significant difference (Desaire et al. 2023), though even among those that do find a difference, there is little consensus regarding whether LLMs produce longer sentences than humans or vice versa. De Mattei et al. (2021) find that their analyzed LLM produced less coherent sentences with longer sentence lengths.

### 4.2.3 Distribution of Parts-of-Speech

A number of studies report findings that pertain to differences in the use or distribution of parts-of-speech. The reported patterns are quite varied, though some overlap can be found. For instance, a number of authors find that AIGT contains more nouns, adpositions, determiners, and coordinating conjunctions, while exhibiting less adverbs in comparison with HWT (Chong et al. 2023; Guo et al. 2023; Liao et al. 2023). Similarly, two studies discover that text generated by LLMs also contains a higher degree of nominalization (e.g. it is more common to find nouns that are derived from other word classes in AIGT, such as *announce - announcement*) (Herbold et al. 2023; Reinhart et al., 2024). AIGT is also reported to contain fewer proper nouns than HWT (Chong et al. 2023; Desaire et al. 2023; Petukhova et al. 2024) and fewer first-person personal pronouns (Mitrović et al. 2023). Interestingly, Muñoz-Ortiz et al. (2024) find that male personal pronouns were more prominent in AIGT. In the same study, they also show that LLMs are more inclined to use symbols and numbers, while producing fewer adjectives, which signals an objective style. Conversely, Rad et al. (2024) find no evidence for differences in the use of part-of-speech patterns between the two types of text.

### 4.2.4 Distribution of Syntactic Structures

In terms of the distribution of syntactic structures, AIGT contains more auxiliary and copula relation types, while containing less adjectival and adverbial modifier relations (Liao et al. 2023; Muñoz-Ortiz et al. 2024). Liao et al. (2023) additionally report a higher prevalence of determiner, conjunct, coordination, and direct object relations in AIGT, while noting that numeral modifiers are less common, which contrasts with Muñoz-Ortiz et al. (2024), who claim that numeral modifiers are more commonly used in AIGT.

### 4.2.5 Order of Sentence Constituents

Simon et al. (2023) focus on the order of sentence constituents and describe that AIGT in English and Spanish very consistently features the canonical Subject-Verb-Object ordering, while a noticeably higher variation of basic sentence constituent orderings can be found in texts produced by human authors.

### 4.2.6 N-Gram Analysis

Two studies report that AIGT tends to contain higher frequencies for some word n-grams (when n equals 3 or more) compared to HWT (André et al. 2023; Merrill et al. 2024). Related to this, Shaib et al. (2024) find that large language models tend to repeat longer part-of-speech sequences more frequently than human authors. McCoy et al. (2023) report that, for values of n higher than 6, LLMs consistently produce more novel[5] word n-grams than humans, while for lower values of n they report a tendency to produce less novel word n-grams.

### 4.2.7 Miscellaneous

Discourse markers seem to be less frequent in AIGT than in HWT, and the discourse markers used in AIGT tend to be more repetitive (Herbold et al. 2023; Simon et al. 2023). AIGT also contains less modal expressions and epistemic markers than HWT (Herbold et al. 2023). Simon et al. (2023) find that AIGT contains less comparative and superlative structures than HWT. AIGT also produces lower readability scores than HWT (Markowitz et al. 2023). Yildiz Durak et al. (2025) discover that AIGT tends to contain less words in the singular than HWT. Reinhart et al. (2024) additionally report that AIGT often exhibits more phrasal coordination, more present participial clauses, and more sentences with *that*-clauses as subjects. Strübbe et al. (2025) also mention a higher occurrence of coordination[6] in AIGT compared to HWT. Wu (2025) finds that AIGT often relies heavily upon cohesive devices and formulaic syntactic constructions. Several studies also mention frequencies for certain part-of-speech bigrams and the rate of function words as useful grammatical features that lead to good classification performance for the AIGT detection systems that they develop (Luo et al. 2023; Petukhova et al. 2024; Zaitsu and Jin 2023).

## 4.3 Other

In addition to the findings above, authors also describe results connected to other miscellaneous areas of linguistic description.

### 4.3.1 Personal References and Named Entities

When analyzing the differences between tweets produced by humans and AI-generated tweets, Chong et al. (2023) find that AI-generated tweets less often contain URLs and mentions of other users than human tweets. The same study also looks at named entity patterns. Personal named entities were found to appear more frequently in AI-generated tweets, while named entities related to dates appear less frequently. Conversely, another study finds that AIGT contains less personal references (Goulart et al. 2024).

### 4.3.2 Stylistic Analysis

---

[5] The authors consider an n-gram produced by an LLM as novel if it does not occur in the training data of the model. For the human baseline, the authors consider an n-gram in the human-written dataset as novel if it is not repeated in the same dataset.
[6] The authors explicitly refer to structures that involve coordination as *enumeration*. It is not clear from the article whether their use of this term is equivalent to the usual use of the term coordination in linguistic literature.

Some studies provide results which focus more on a particular stylistic quality of the text output of LLMs. Two studies report that AIGT contains a more formal style than HWT (Kabir et al. 2024; Zhang and Crostwaithe 2025). Markowitz et al. (2023) find that AIGT tends to be more analytic and use more descriptive language than HWT. AIGT is found to be less narrative than HWT and is directed more towards conveying information rather than signaling involvement (Sardinha 2024). Goulart et al. (2024) similarly observe that AIGT features a less involved, more impersonal style than HWT. One study describes a qualitative assessment of AIGT, in which human evaluators attempted to determine whether a text was produced by an LLM or a human author. The participants most commonly reported the following characteristics as determining factors in their choice: "the general ease or difficulty of reading", "the tone or voice [...] seems familiar, humanlike, or personal in nature", "a particular phrase or sequence of words seems well produced or constructed" (Casal and Kessler 2023: 9). Two studies report that AIGT tends to contain content that is more abstract than HWT (Casal and Kessler 2023; Liao et al. 2023). However, Sardinha (2024) discovers that how abstract the AIGT is may depend greatly on the type of text the LLM is tasked to produce (see Section 4.5).

**4.3.3 Sentiment Analysis and Emotional Content**

Several studies have also addressed the somewhat related concepts of sentiment analysis and presence of emotional content. AIGT is consistently found to contain more neutral sentiment than HWT (Chong et al. 2023; Guo et al. 2023; Liao et al. 2023), while study results are not consistent in regard to the prevalence of positive and negative sentiment (Chong et al. 2023; Kabir et al. 2024; Liao et al. 2023). As for the presence of emotional content, in their analysis of AI-generated news articles, Muñoz-Ortiz et al. (2024) find that AIGT contains less negative emotions like fear and anger. On the other hand, in an analysis of AI-generated hotel reviews, Markowitz et al. (2023) discover that AI-generated text is in general more emotional than HWT.

# 5 Methodological Considerations

In this section we present the results reported by the obtained articles in relation to several unique methodological considerations related to the analysis of AIGT. In particular, we focus on study results that uncover the differences between texts produced by different language models (Section 5.1), results that pertain to a specific genre (Section 5.2), a specific language (Section 5.3), and a specific approach to prompting (Section 5.4).

## 5.1 Differences Between Models

Some of the reviewed works also compared texts produced by different LLMs to expose potential differences between models. Herbold et al. (2023) find that, while lexical diversity is higher in HWT compared to AIGT produced by GPT-3.5, it is also lower compared to text generated by GPT-4. Liebe et al. (2023) discover that while texts produced by GPT-2 can be more easily detected using a classifier based on character trigrams, a classifier directed at detecting GPT-3 performs better if it utilizes syntactic trigrams based on syntactic trees. Rosenfeld and Lazebnik (2024) find that Bard gives shorter answers to questions, has a smaller vocabulary size, and a higher density[7] than GPT-3.5 and GPT-4. They also observe that, compared to GPT-3.5, GPT-4 gives shorter answers and exhibits a larger vocabulary size and density. Text generated by Bard also seems to contain more low-frequency parts-of-speech and more high-frequency dependency relations than both GPT models. Muñoz-Ortiz et al. (2024) observe that the amount of negative emotions present in text generated by Llama models increased as the model size in parameters grew (from 7B to 65B). Text generated by BloomZ was found to be concise and repetitive, while Qwen2.5 in comparison was found to be much closer in overall linguistic style to HWT (Culda et al. 2025). Reinhart et al. (2024) observe that the GPT-4o and GPT-4o-mini models use downtoners (i.e. expressions such as *barely* and *nearly*) very often and refrain from using clausal coordination, while variants of Llama 3 exhibit the opposite, avoiding the use of downtoners and use more clausal coordination. When comparing text generated by GPT-4o, BingAI, and Gemini, Yildiz Durak et al. (2025) find that AIGT generated by BingAI has shorter sentences, while AIGT generated by Gemini features a higher number of words and punctuation compared to the other models.

## 5.2 Differences Between Genres

Some authors also explore whether AIGT produces different kinds of linguistic output in different genres, though, among all the analyzed articles in this survey, only one such study (Sardinha 2024) deals with genre comparison in an extensive manner. In this study, Sardinha (2024) analyzes AI-generated essays, academic texts, news texts, and conversations. For AI-generated academic texts, they find that they rarely contain various narrative elements, lack explicit references and contain more persuasive language than human-written academic texts. With essays, they discover that AIGT is very information-dense, exhibits less involvement and integration, and contains less persuasive language than HWT. They find AI-generated news articles to exhibit less involvement and less narrative elements than human-written news articles. Lastly, with conversations, they also note that AI-generated conversations show less involvement and integration and contain less narrative and persuasive elements than in human conversations, while the language employed is more abstract than in human conversations.

Petukhova et al. (2024) find that AI-generated Wikipedia entries differ quite substantially from the Wikipedia articles produced by human authors, while also noting that the differences between Wikipedia articles produced by different models were quite large. In contrast, texts from the WikiHow domain included in this study do not show such drastic differences. Regardless, the authors do not elaborate on these differences in more detail.

## 5.3 Differences Between Languages

Although the vast majority of the inspected papers focus on English as the main language of the analyzed texts, the studies that do focus on other languages as well occasionally reveal some important findings.

The prevalence of grammatical errors in AIGT seems to depend largely on the language of the generated text. English AIGT contains less grammatical errors than HWT (André et al. 2023), while one study on AI-generated clinical text in Japanese finds that generated content often contains grammatical mistakes, uses erroneous writing systems, and even uses the wrong language for some expressions (Yanagita et al. 2024).

---

[7] The authors define their *density* score in a similar way that the most common measures of lexical diversity are calculated, such as the type-token-ratio.

When comparing question answering texts in Chinese and English, Guo et al. (2023) note that the human-produced answers are shorter and contain a larger vocabulary size, and that this tendency is more pronounced in Chinese texts than in English. However, the tendency can be found in both languages to some extent.

**5.4 Differences Between Prompting Strategies**

None of the 9 papers that use more than one type of prompt to obtain AI-generated data for their experiments focus specifically on comparing the linguistic content of the output produced by different prompts. Two studies note that the AIGT vs. HWT classification systems they developed perform notably worse on AIGT which was generated using a revised prompt. In contrast to the basic prompt, which simply instructed the model to produce text based on a title, this revised prompt instructed the model to rephrase existing human-written text (Kim and Desaire 2024; Mindner et al. 2023). This suggests that providing extant human-written text to the model in the prompt and instructing the model to rephrase it produces text that is closer to human-written text in terms of linguistic style. Still, the two papers do not explore in further detail the specific linguistic differences in the output produced by the two different prompting strategies.

# 6 Discussion

Many of the findings presented in the previous section represent important linguistic aspects of AIGT, however some discoveries stand out as particularly compelling, especially those that are supported by several studies. At the grammatical level, AIGT most notably contains more nouns, determiners, and adpositions, which correspond to parts-of-speech that often form the core of nominal phrases. This seems to be closely tied to the observation that AIGT contains more nominalization than HWT. LLMs appear to generate language that has more characteristics of formal registers, placing much emphasis on nominal phrases and avoiding excessive use of adjectives and adverbs, which signal more involvement from the side of the speaker. To a degree, this also connects to the observation that AIGT is more likely to contain more neutral sentiment than HWT, as this also suggests a more formal and impersonal style of writing.

In terms of lexicon, a great number of studies reach the same conclusion that AIGT is generally less lexically diverse than text written by humans. An interesting link can be observed between this and the observations that AIGT exhibits a smaller vocabulary size, a higher rate of repetition, and a higher frequency of use for certain n-grams of higher orders in comparison to HWT. This suggests that these generative AI systems possess a strong tendency to use certain words, expressions, or even sentence patterns more frequently. These patterns may be learned by the model during the training process as an effective method to reliably generate natural-sounding text without the need of demonstrating a large vocabulary diversity. The fact that these tendencies can be detected only when the texts are assessed using detailed methods of linguistic analysis suggests that humans are not usually mindful of such obscure patterns of repetition but nevertheless still avoid them when producing language.

In addition to findings that are corroborated by several different works, there are also points of disagreement where authors of different studies reach different conclusions. Points where there is especially little consensus include word length, sentence length, the distribution of certain syntactic structures, named entity references, sentiment, and emotional content. As pointed out by Sardinha (2024), differences in the degree of abstractness in AIGT only become apparent when taking into account the text genre, so future studies should take special care to account for such potential influencing factors when revisiting these points of disagreement.

Despite the relatively large amount of work that has been done to identify AIGT linguistic characteristics, it is not clear whether the discoveries made so far truly apply to text produced by LLMs in general or are they restricted to only certain situations. For instance, in the extant body of research there is a notable lack of diversity when it comes to the types of models used, as only approximately 43% of the analyzed articles employ a model other than GPT-3.5 to conduct their experiments. This disproportionately high use of a single model could mean that the majority of the findings reported in this survey could apply primarily to GPT-3.5, as it is not clear whether what holds for AIGT produced by one model should apply also to other models.

We present a number of studies that report on linguistic differences between the output of several models, however, as we note above, these studies are relatively few among the ones included here. In fact, of the 44 articles only 15 (approx. 34%) report that they use models from more than one model family to obtain experiment data. As the comparison between the output of different models in Section 4.4 suggests, the differences between models can be quite large, thus researchers should include more than one type of large language model in their experiments.

Additionally, most research so far has been conducted on closed models, primarily developed by OpenAI, while open models, which are publicly accessible and for which the training process and internal architecture are often more transparent, are featured in relatively few works. Transparency is an important part of the scientific process and, if this trend of experiments on closed models persists, a situation could arise in which a lot is known about the output of the best performing models, but next to nothing about their development and training phases, and vice versa.

Generally, the more formal and rigid textual genres are more commonly used as bases for experiments in our analyzed studies than the more everyday types of texts, as is evident in more articles focusing on scientific texts, news articles, and essays than social media texts and hotel or restaurant reviews. One possible explanation for such a trend is availability of data, since scientific texts and news articles are considerably easier to obtain from online sources than more informal texts.

Far more research has been conducted for English than for any other language, reflecting the enduring dominance of English over other less well-resourced languages in the field of natural language processing. This is concerning as some modern LLMs achieve impressive results also on some languages with less available resources and representation, as is evident, for instance, in the evaluation results for Welsh, Swahili, and Latvian provided in the technical report for GPT-4 (OpenAI 2024). If LLM-based technology gains an even more widespread presence on a global scale, research on the characteristics of AIGT in other languages should definitely not be overlooked.

Researchers seem to be quite aware of the importance of reporting the form of the prompt used to obtain data for research, as all but three out of the 44 articles give some description of the employed prompts, though not all provide the text that was used to prompt the model verbatim. As mentioned before, a total of 9 studies also report that they use more than one concrete prompt wording to obtain their AI-generated data. The practice of using more than one prompt wording to obtain experiment data is quite crucial, as the sensitivity of LLM output to the provided prompt could lead to text output that contains different linguistic features depending on the specific input used during generation. Researchers should therefore take care to evaluate the linguistic content of the AIGT produced using a range of different prompts so as to increase the robustness of their data.

# 7 Conclusion

This survey paper has presented the current situation of what is known about the linguistic characteristics of text generated by large language models. The range of possible linguistic investigations that can be performed on AI-generated text is quite extensive and there are many areas that remain to be explored. However, many intriguing discoveries have been made. Among other things, it has been shown that AIGT is more likely to contain certain parts-of-speech such as nouns, determiners, and adpositions, while it is less likely to contain adjectives and adverbs. This suggests that a more formal and impersonal style is favored by LLMs. At the same time, lexical diversity and vocabulary size are generally lower in AIGT compared to HWT and instances of repetition are more frequent.

Future studies should be mindful of several drawbacks of the current body of research and methodological considerations that should be taken into account.:

1. As most research so far has been conducted on English text generated using the GPT-3.5 model, scholars should diversify their choice of models in future experiments, focusing also on less-used open models for which the information about the training process is made public and transparent.
2. Languages other than English, including low-resource languages, should also be given more attention in future research, as LLMs perform increasingly well also on non-English generation tasks.
3. More attention should be devoted to obtaining generated texts for linguistic analysis using more than one prompt wording, as these can significantly influence output.
4. Lastly, future studies should aim to uncover not only what linguistic features set AIGT as a whole apart from HWT, but also what differences exist between texts generated by different generative models, different genres of text, texts in different languages, and texts generated using different prompt wordings.

## Acknowledgements

The work described in this paper was made possible by the Young researchers program, the Language Resources and Technologies for Slovene research program (project ID: P6-0411), the PoVeJMo research project (Adaptive Natural Language Processing with the Help of Large Language Models), and the Large Language Models for Digital Humanities project (Grant GC-0002), all financed by the Slovenian Research and Innovation Agency (ARIS).

29. Meskó, Bertalan. 2023. Prompt engineering as an important emerging skill for medical professionals: Tutorial. *Journal of Medical Internet Research* 25. https://doi.org/10.2196/50638.

30. Mikros, George, Athanasios Koursaris, Dimitrios Bilianos & George Markopoulos. 2023. AI-writing detection using an ensemble of transformers and stylometric features. In *IberLEF 2023*, Jaén, Spain. https://ceur-ws.org/Vol-3496/autextification-paper9.pdf.

31. Mindner, Lorenz, Tim Schlippe & Kristina Schaaff. 2023. Classification of human-and AI-generated texts: Investigating features for ChatGPT. In *International Conference on Artificial Intelligence in Education Technology*, Singapore. https://doi.org/10.1007/978-981-99-7947-9_12.

32. Mitrović, Sandra, Davide Andreoletti & Omran Ayoub. 2023. ChatGPT or human? Detect and explain. Explaining decisions of machine learning model for detecting short ChatGPT-generated text. *arXiv preprint*. https://doi.org/10.48550/arXiv.2301.13852.

33. Mizrahi, Moran, Guy Kaplan, Dan Malkin, Rotem Dror, Dafna Shahaf & Gabriel Stanovsky. 2024. State of what art? A call for multi-prompt LLM evaluation. *Transactions of the Association for Computational Linguistics* 12. 933–949. https://doi.org/10.1162/tacl_a_00681.

34. Muñoz-Ortiz, Alberto, Carlos Gómez-Rodríguez & David Vilares. 2024. Contrasting linguistic patterns in human and LLM-generated news text. *Artificial Intelligence Review* 57(10). 265. https://doi.org/10.1007/s10462-024-10903-2.

35. Nguyen, Trung T., Amartya Hatua & Andrew H. Sung. 2023. How to detect AI-generated texts? In *2023 IEEE 14th Annual Ubiquitous Computing, Electronics & Mobile Communication Conference (UEMCON)*. https://doi.org/10.1109/UEMCON59035.2023.10316132.

36. Nkhobo, Tlatso Ishmael & Chaka Chaka. 2023. Student-written versus ChatGPT-generated discursive essays: A comparative Coh-Metrix analysis of lexical diversity, syntactic complexity, and referential cohesion. *International Journal of Education and Development using Information and Communication Technology* 19(3). 69–84. https://www.researchgate.net/publication/377020171_Student-Written_Versus_ChatGPT-Generated_Discursive_Essays_A_Comparative_Coh-Metrix_Analysis_of_Lexical_Diversity_Syntactic_Complexity_and_Referential_Cohesion.

37. OpenAI, Josh Achiam, Steven Adler, Sandhini Agarwal, Lama Ahmad, Ilge Akkaya, Florencia Leoni Aleman, Diogo Almeida, Janko Altenschmidt, Sam Altman, Shyamal Anadkat, Red Avila, Igor Babuschkin, Suchir Balaji, Valerie Balcom, Paul Baltescu, Haiming Bao, Mohammad Bavarian, Jeff Belgum, Irwan Bello, Jake Berdine, Gabriel Bernadett-Shapiro, Christopher Berner, Lenny Bogdonoff, Oleg Boiko, Madelaine Boyd, Anna-Luisa Brakman, Greg Brockman, Tim Brooks, Miles Brundage, Kevin Button, Trevor Cai, Rosie Campbell, Andrew Cann, Brittany Carey, Chelsea Carlson, Rory Carmichael, Brooke Chan, Che Chang, Fotis Chantzis, Derek Chen, Sully Chen, Ruby Chen, Jason Chen, Mark Chen, Ben Chess, Chester Cho, Casey Chu, Hyung Won Chung, Dave Cummings, Jeremiah Currier, Yunxing Dai, Cory Decareaux, Thomas Degry, Noah Deutsch, Damien Deville, Arka Dhar, David Dohan, Steve Dowling, Sheila Dunning, Adrien Ecoffet, Atty Eleti, Tyna Eloundou, David Farhi, Liam Fedus, Niko Felix, Simón Posada Fishman, Juston Forte, Isabella Fulford, Leo Gao, Elie Georges, Christian Gibson, Vik Goel,

# Appendix A

Table containing the categorization of every article included in the survey.

| Article | Levels of linguistic analysis | Model type used | Type/genre of generated texts | Approach to prompting | Languages used | Features analyzed |
|---|---|---|---|---|---|---|
| (Herbold et al., 2023) | Grammar, Lexicon, Other | GPT-3.5, GPT-4 | Essays | Single prompt | English | Syntactic complexity, POS distribution, Miscellaneous grammatical features |
| (Muñoz-Ortiz et al., 2024) | Grammar, Lexicon, Other | LLaMA 1 | News articles | Single prompt | English | Lexical diversity, POS distribution, Syntactic structure distribution, Emotional content |
| (Markowitz et al., 2023) | Grammar, Other | GPT-3.5 | Restaurant/hotel reviews | Single prompt | English | Miscellaneous grammatical features, Stylistic features, Emotional content |
| (Zindela, 2023) | Grammar, Lexicon, Other | GPT-3.5 | Essays | Single prompt | English | Lexical diversity, Sentence length |
| (Nkhobo & Chaka, 2023) | Grammar, Lexicon, Other | GPT-3.5 | Essays | Single prompt | English | Syntactic complexity |
| (Casal & Kessler, 2023) | Other | GPT-4 | Scientific texts | Single prompt | English | Stylistic features |
| (Cai et al., 2023) | Grammar, Lexicon, Other | GPT-3.5 | Question answering | Single prompt | English | Word length |
| (Mindner et al., 2023) | Grammar, Other | GPT-3.5 | Wikipedia articles | Multiple prompts | English | Miscellaneous lexical features |
| (Zaitsu & Jin, 2023) | Grammar, Other | GPT-3.5, GPT-4 | Scientific texts | Single prompt | Japanese | Miscellaneous grammatical features |
| (Guo et al., 2023) | Grammar, Lexicon, Other | GPT-3.5 | Question answering | Single prompt | English, Chinese | Lexical diversity, Syntactic complexity, Sentence length, POS distribution, Sentiment analysis |
| (Simón et al., 2023) | Grammar, Lexicon, Other | GPT-3, BLOOM | News articles, Social media posts, Restaurant/hotel reviews, Other | Single prompt | English, Spanish | Repetition, Miscellaneous lexical features, Sentence constituent ordering, Miscellaneous grammatical features |
| (Liebe et al., 2023) | Grammar, Lexicon, Other | GPT-2, GPT-3 | News articles | Single prompt | English | N-grams |
| (Chong et al., 2023) | Grammar, Lexicon, Other | Unspecified | Social media posts | Prompt not specified | English | POS distribution, Named entities, Sentiment analysis |
| (Shah et al., 2023) | Grammar, Lexicon, Other | Orca-Mini, GPT4All-J | Wikipedia articles | Single prompt | English | Lexical diversity |
| (Sardinha, 2024) | Other | GPT-3.5 | Essays, News articles, Scientific texts, Other | Single prompt | English | Stylistic features |
| (Nguyen et al., 2023) | Grammar, Lexicon, Other | GPT-3.5 | News articles, Wikipedia articles | Single prompt | English | Miscellaneous lexical features |
| (Luo et al., 2023) | Grammar, Lexicon, Other | GPT-3 | Restaurant/hotel reviews | Single prompt | English | Miscellaneous grammatical features |
| (Liao et al., 2023) | Grammar, Lexicon, Other | GPT-3.5 | Scientific texts | Multiple prompts | English | Lexical diversity, Syntactic complexity, POS distribution, Syntactic structure distribution, Stylistic features, Sentiment analysis |
| (Liu et al., 2023) | Grammar, Lexicon, Other | GPT-2, GPT-3, GPT-3.5 | Essays | Multiple prompts | English | Lexical diversity, Syntactic complexity, Sentence length |
| (De Mattei et al., 2021) | Grammar, Lexicon, Other | GPT-2 | Unspecified | Prompt not specified | Italian | Sentence length |
| (Seals & Shalin, 2023) | Other | GPT-3.5 | Scientific texts | Multiple prompts | English | Lexical diversity, Syntactic complexity |
| (Desaire et al., 2023) | Lexicon, Other | GPT-3.5 | Scientific texts | Single prompt | English | Specific vocabulary, Miscellaneous lexical features, Sentence length, POS distribution |
| (Rosenfeld & Lazebnik, 2024) | Grammar, Lexicon | GPT-3.5, GPT-4, Bard | Question answering | Single prompt | English | Lexical diversity, Sentence length |
| (André et al., 2023) | Grammar, Lexicon | GPT-3.5 | Scientific texts | Multiple prompts | English | Lexical diversity, Word length, N-grams |
| (Kim & Desaire, 2024) | Grammar, Lexicon | GPT-3.5 | News articles | Multiple prompts | English | Specific vocabulary |
| (Rad et al., 2024) | Grammar | GPT-3, GPT-3.5, GPT-4, BLOOM, Cohere, Dolly | News articles, Wikipedia articles, Scientific texts, Question answering | Multiple prompts | English, Chinese, Arabic, Indonesian, Russian, Urdu | POS distribution |
| (Mikros et al., 2023) | Grammar, Lexicon | GPT-3, BLOOM | Social media posts, Other | Single prompt | English | Miscellaneous lexical features |
| (Yanagita et al., 2024) | Grammar, Lexicon, Other | GPT-4 | Scientific texts | Single prompt | Japanese | Repetition, Miscellaneous lexical features |
| (Fujiwara, 2024) | Grammar, Lexicon | GPT-3.5 | Essays | Single prompt | English | Specific vocabulary |

| Reference | Linguistic aspects | Models | Text types | Prompts | Languages | Features |
|---|---|---|---|---|---|---|
| (Mitrović et al., 2023) | Grammar, Lexicon | GPT-3.5 | Restaurant/hotel reviews | Multiple prompts | English | Specific vocabulary, POS distribution |
| (Petukhova et al., 2024) | Grammar, Lexicon, Other | GPT-3, GPT-3.5, GPT-4, BLOOM, Cohere, Dolly | News articles, Wikipedia articles, Scientific texts, Question answering | Multiple prompts | English | POS distribution, Miscellaneous grammatical features |
| (Shaib et al., 2024) | Grammar | OLMo, Mistral, Llama 2, Llama 3, GPT-4o, Alpaca | Scientific texts, Social media posts, Other | Single prompt | English | N-grams |
| (Merrill et al., 2024) | Grammar, Lexicon | Pythia | Other | Single prompt | English | N-grams |
| (McCoy et al., 2023) | Grammar, Lexicon | Original Transformer | Wikipedia articles, Other | Single prompt | English | N-grams |
| (Culda et al., 2025) | Grammar, Lexicon | BLOOM, GPT-Neo, Qwen2.5 | Wikipedia articles, Scientific texts, Social media posts | Single prompt | English | Lexical diversity, Syntactic complexity, Sentence length |
| (Reinhart et al., 2024) | Grammar, Lexicon | Llama 3, GPT-4o | News articles, Scientific texts, Social media posts, Other | Single prompt | English | POS distribution, Miscellaneous grammatical features |
| (Strübbe et al., 2025) | Grammar | GPT-4o, OpenAI o1, GPT-3.5, GPT-4 | News articles, Wikipedia articles, Other | Single prompt | English, French, German | Miscellaneous grammatical features |
| (Goulart et al., 2024) | Grammar, Lexicon | GPT-3.5 | Essays, Scientific texts | Single prompt | English | Named entities, Stylistic features |
| (Yildiz Durak et al., 2025) | Grammar, Lexicon | GPT-4o, BingAI, Gemini | Essays | Single prompt | Unspecified | Lexical diversity, Word length, Miscellaneous grammatical features |
| (Wu, 2025) | Grammar, Lexicon, Other | GPT-4 | Essays | Single prompt | English | Miscellaneous grammatical features |
| (Arcenal et al., 2024) | Grammar, Lexicon, Other | GPT-4 | Social media posts | Prompt not specified | English | Repetition |
| (Zhang & Crosthwaite, 2025) | Grammar, Lexicon | GPT-3.5 | Essays | Single prompt | English | Lexical diversity, Stylistic features |
| (Wan, 2024) | Grammar, Lexicon | GPT-4 | Restaurant/hotel reviews | Single prompt | English | Specific vocabulary |
| (Kabir et al., 2024) | Grammar, Lexicon, Other | GPT-3.5 | Question answering | Single prompt | English | Stylistic features, Sentiment analysis |